Bruno Gonçalves

David Sánchez

# Learning about Spanish dialects through Twitter

**Abstract:** This paper maps the large-scale variation of the Spanish language by employing a corpus based on geographically tagged Twitter messages. Lexical dialects are extracted from an analysis of variants of tens of concepts. The resulting maps show linguistic variation on an unprecedented scale across the globe. We discuss the properties of the main dialects within a machine learning approach and find that varieties spoken in urban areas have an international character in contrast to country areas where dialects show a more regional uniformity.

**Resumen:** En este trabajo, cartografiamos la variación a gran escala del idioma español usando un corpus basado en mensajes geolocalizados de Twitter. Se extraen las formas dialectales léxicas a partir del análisis de decenas de variantes. Los mapas resultantes muestran una variación lingüística en todo el planeta con una escala que no tiene precedentes. Examinamos las propiedades de los principales dialécticos empleando técnicas de aprendizaje automático y hallamos que las variedades habladas en áreas urbanas poseen un carácter internacional, a diferencia de las zonas rurales, donde los dialectos presentan una uniformidad más regional.

**Keywords:** Spanish variation, Twitter, geolocation, lexical dialectology

**Palabras clave:** variación del español, Twitter, geolocalización, dialectología léxica

## 1. Introduction[1]

As all social animals, humans depend on communication to live, grow, and thrive and as society grows ever more complex so does the need to communicate effectively. The evolutionary answer to this need was the development of languages, first in spoken and later in written form.

---

[1] BG thanks the Moore and Sloan Foundations for support as part of the Moore-Sloan Data Science Environment at NYU.

Language allows us to clearly express our thoughts and feelings, transmit knowledge, coordinate large groups of individuals and ultimately achieve our societal and individual goals. For a given language to be able to play its fundamental social role it must continuously evolve and adapt to serve the needs of its population. As a result, new forms, registers, dialects and even whole languages emerge and fade from existence.

Over the years, linguists have strived to study and analyze all aspects of variation and change in language (Chambers / Trudgill 1998), both oral (Labov / Sharon / Boberg 2005) and written (Bauer 2004) using surveys, interviews, specially crafted corpora, etc. Studies that relied predominantly on written records have focused on more formal types of language use, leaving the vernacular as a little explored and understood domain. The recent massification of online social networking and microblogging services resulted in an unprecedented wealth of written content produced by large swaths of the population in many different contexts. Not surprisingly, these new Internet corpora have attracted the attention of linguists of diverse backgrounds and opened the doors to new and innovative studies on how language use varies both geographically and over time (Nguyen *et al*. 2015) in different languages. Twitter is an excellent example. For instance, using this tool we (Gonçalves / Sánchez 2014) find two global varieties of Spanish; Einsenstein *et al.* (2014) propose a latent variable model for English geographic lexical diffusion and change; Doyle (2014) discusses the differences between Twitter-based linguistic maps and results from more traditional approaches; Ibrahim / Abdou / Gheith (2014) use standard Arabic and Egyptian dialectal Arabic tweets in a sentiment analysis; Kulkarni / Perozzi / Skiena (2015) examine semantic and syntactic variation of English with a massive online dataset; Estrada Arráez / de Benito Moreno (2016) investigate language innovations in online social networks.

In this work, we build on our previous efforts (Gonçalves / Sánchez 2014) and use Twitter to empirically define Spanish geographically consistent dialects. Based on a machine learning analysis we found that Spanish is globally split into two large lexical clusters. Cluster α corresponds to the speech used in largely populated areas (an international variety) while cluster β is mostly encountered in rural areas and is thus related to local varieties. Here, using a greatly expanded set of tweets and an independent list of words we verify the existence of these superdialects and recompute the dialect isoglosses.

## 2. Methods

Using the Twitter Gardenhose we collected an unbiased sample of all tweets produced between May 2010 and June 2015. See Mocanu *et al.* (2013), Gonçalves / Sánchez (2014) and Ronen *et al.* (2014) for further results on Twitter datasets. From these, we selected the subset of tweets containing geolocation information and used the Google's Compact Language Detection (McCandless 2012) library to identify all tweets written in Spanish. The resulting dataset contains 106 million geolocated tweets written in Spanish. As shown in Figure 1, the overwhelming majority of Spanish tweets are located in Spain and Spanish speaking Latin-American countries (Moreno Fernández / Otero Roth 2007) with the remaining tweets being attributable to regions with large expat communities or large tourist attractions.

**Figure 1: Geographical distribution of Spanish tweets in our dataset.**

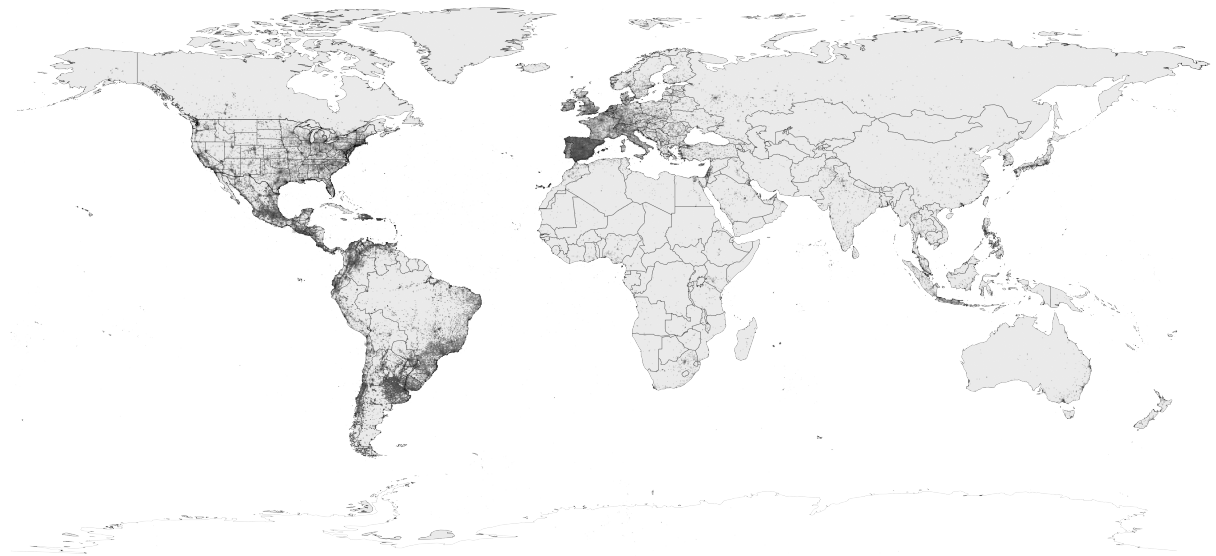

The next step is to define a corpus. We focus on word choice (lexical differences) because this level of language is more easily accessed in our database. The advantage of using lexical variation is methodological as studying phonetic variation (Penny 2000) in written corpora poses many practical challenges. See, e.g., Cahuzac (1980) for a lexical analysis that leads to clear divisions of Hispanic America. Our corpus is built from the Varilex database (Ueda / Takagaki /

Ruiz Tinoco 1993). Varilex contains thousands of words and expressions grouped by concepts. The different correspondences for a given concept are registered in many Spanish-speaking locations. We selected a list of 46 common concepts containing 331 lexical features (the full list is shown in the Appendix below) and isolated 4 million instances in which these words appeared in our dataset of 106 million geolocated tweets in Spanish. The selection of these concepts is aimed at examining only geographical distinctions, whilst also minimizing semantic ambiguities and register variations. Each instance was mapped to a 25"x25" geographical cell (we have in total 3629 non-empty cells) for which the dominant word for each concept was determined. Maps for each concept were generated by drawing a circle centered at each cell with a color given by the dominant word and an area that scales with how many times that word was observed in that cell.

Finally, we built a cell x word matrix, *M*, where element $M_{cw}$ is 1 if word *w* is the dominant term for a given concept in cell *c* and 0, otherwise. This matrix summarizes all the relevant lexico-geographical information that is necessary to fully characterize the spatial variation of the Spanish language.

## 3. Results and discussion

Figure 2 shows the two maps corresponding to the concepts 'to miss [someone]' (left) and 'cold' (right). From this figure it becomes immediately clear that some forms are clearly dominant in some regions only to be supplanted by other forms elsewhere. For example, *echar de menos* is the dominant form in European Spanish while *extrañar* is the more common form in the Spanish spoken in America. On the other hand, the geographical distribution of the words corresponding to 'cold' is much more varied with *resfrío* being common in the Río de la Plata region, especially around Buenos Aires, *gripa* being dominant in Mexico and Colombia and several other forms coexisting in the Iberian Peninsula.

**Figure 2: Geographical distribution of words for the concepts 'to miss [someone]' (left) and 'cold, influenza' (right).**

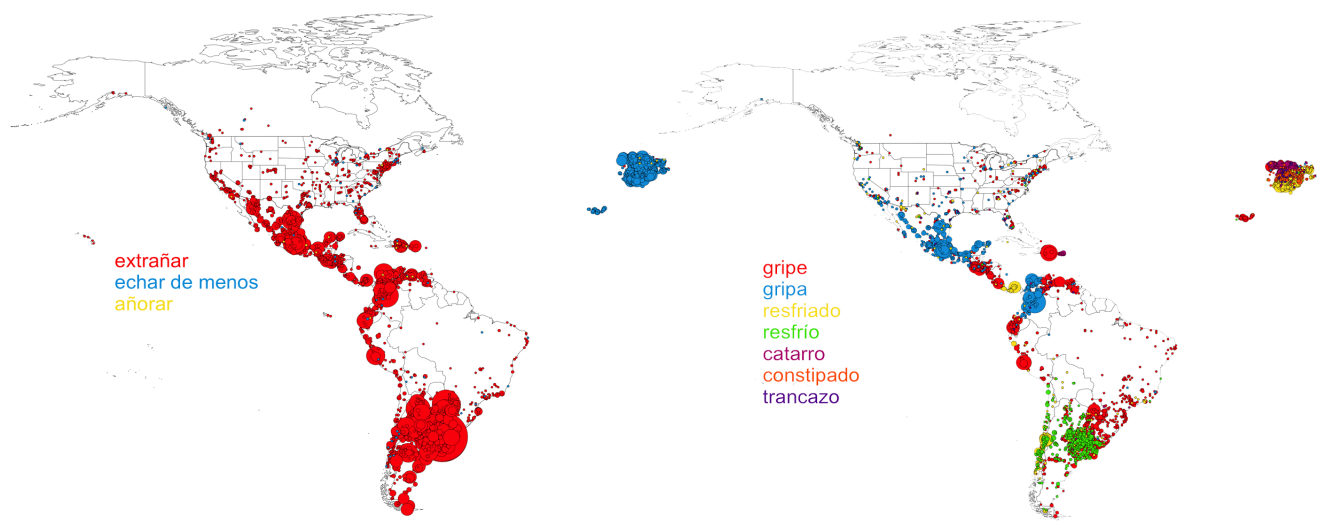

It is worth highlighting the fine spatial detail that we are able to obtain with this approach. In Fig. 3 we zoom in on the geolocations for the words belonging to the concept 'cold' used in Spain. While *resfriado* dominates in Southern Spain the situation is more diverse in the Northern part (Fernández-Ordóñez 2012). Clearly, the final distribution is not only geolectal but can also include register competition. However, we note that our findings are consistent with the results found in Varilex, so the outcome of our methodology is encouraging despite its obvious limitations. Furthermore, in contrast to Varilex our generated maps are not restricted to a few towns but show a geographical continuum for the use of the lexical units. In addition, the maps are based on an automatic analysis of thousands of tweets including the different words.

**Figure 3: Geographical distribution of words for the concept 'cold' in Spain.**

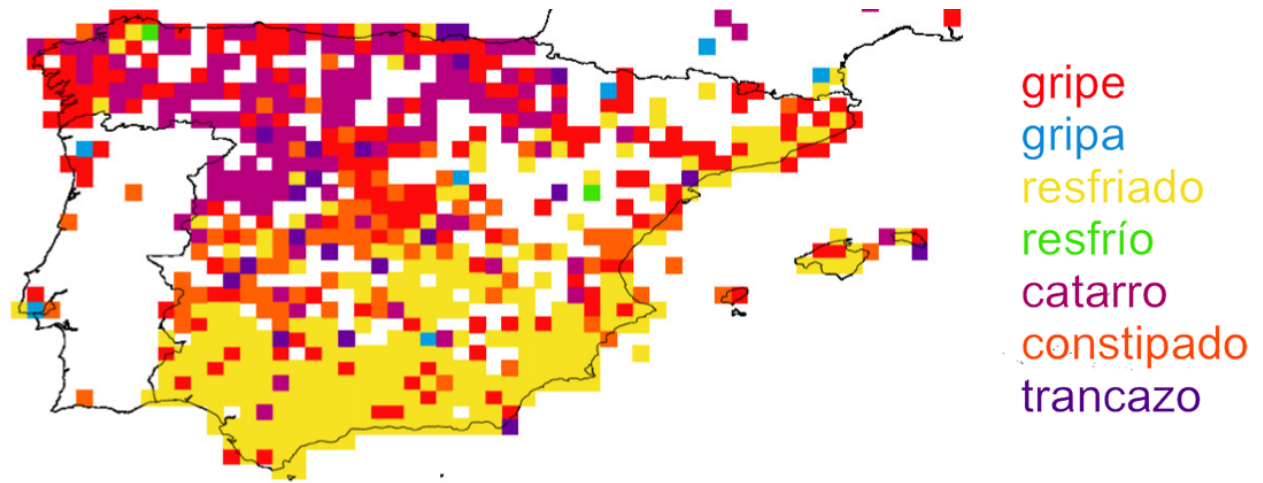

The immense amount of data allows us to address a statistical analysis of our results. Our aim is to unveil the internal structure of the variation for the Spanish used in Twitter. The different dialects and varieties should naturally emerge from the analysis. Many linguistic criteria at different levels have been suggested in order to establish well-defined zones with dialect similarities. Here, we follow a machine learning approach that consists of trying to find regularities in our data (Murphy 2012). The basic idea is to group cells in $K$ clusters that are mathematically similar to each other. This similarity in our data has a linguistic origin and is encoded in $M$. We therefore use a $K$-means algorithm applied to the matrix $M$. However, $M$ is too large. We can obtain a more manageable arrangement by reducing the $M$ size keeping at the same time 95% of the variance (i.e., the meaningful correlations). The difficulty now is to establish an optimal value of K that balances the complexity and a relatively small number of clusters. This value is found from the metric $f(k)$ (Pham / Dimov / Nguyen 2005).

We find that the cells split into two well-defined clusters or macrovarieties, which we represent in the map of Fig. 4 (red dots: superdialect α; blue dots: superdialect β). Surprisingly, the clusters are not localized around a definite region. Both groups are present in all Spanish-speaking countries (except Cuba, for which we have no Twitter data). There have been previous proposals that put forward a bipartition of Spanish into two superdialects. The criterion has been mainly phonetic. For instance, depending on the realization of the implosive /-s/ Fernández Sevilla (1980) and Montes Giraldo (1982) have established two large Spanish varieties: the Southern, atlantic or lowland dialect and the Central, interior or highland dialect. Both superdialects are at the same time present in Europe and America and have no geographical continuity, as in our case. Nevertheless, the nature of our superdialects is entirely different because our analysis is based on the variation of word choice.

Let us examine in closer detail the properties of our two clusters. In the inset of Fig. 4 we plot the statistical distribution of population estimated for each superdialect. Clearly, cluster α has on average more population than cluster β. We confirm this result by checking that many red dots in Fig. 4 correspond to cities and large urban areas (indicated in the figure). We emphasize that this conclusion agrees with our previous work (Gonçalves / Sánchez 2014) but is obtained from a completely different corpus. The extension of the two superdialects slightly differs as compared with our earlier analysis since the Twitter dataset is now larger. In any case, the presence of the two superdialects obtained through independent datasets shows the robustness of our conclusion.

What is the nature of each superdialect? The answer should be found in the influence of cities in the evolution of language. Urban areas have a pivotal role in the globalization process of Spanish (López Morales 2001). The driving force is multiple (mass media, travellers, emigrants, Internet), which tends to a uniformization of the active lexicon eliminating specific words and expressions with a marked regional character. We can assume that the pan-Hispanic interurban variety enjoys social prestige among Twitter users. We have checked that superdialect α presents most of the words in our corpus whereas superdialect β is linguistically more heterogeneous, as expected for rural areas. We assign to superdialect α an international Spanish variety understood and propagated mostly in the main urban centers and likely related to official media.

**Figure 4: After a cluster analysis, cells represented in Fig. 1 fall into two large clusters or superdialects, represented by red (superdialect α) and blue (superdialect β) dots. The inset highlights the population differences between the two clusters.**

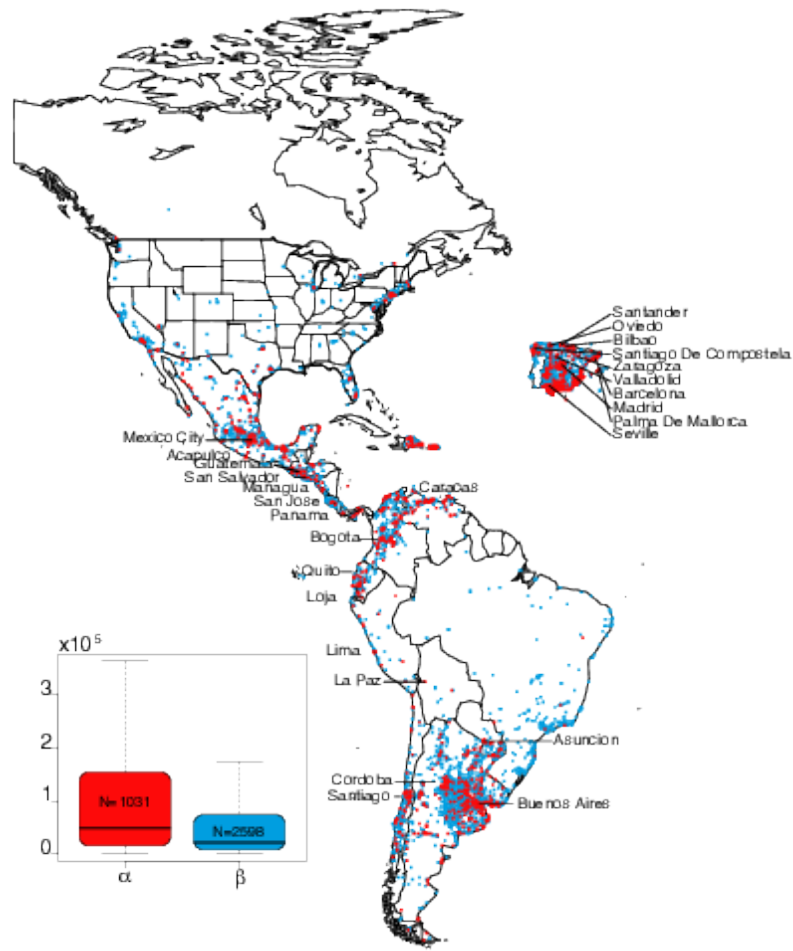

We continue our geographical delineation of the various dialects spoken in Twitter by further subdividing superdialect β. We identify three distinct regional dialects, as shown in Fig 5, corresponding, respectively, to the Iberian Peninsula (in blue), North America, Central America and the Northern part of South America (in orange) and the Southern Cone (in green). Surprisingly, we also find one fourth regional dialect (in yellow) whose extension is less confined and which can be found scattered throughout South America with a predominance in the interior of Argentina, where it coexists with the green dialect, and locations along the Andes mountain range (South, Central and North Andes), where we observe a contact with the orange cluster. These diatopically opposable options are truly dialects understood as regional varieties. The dialect division of Hispanic America is a topic of ongoing debate; for a review, see Moreno Fernández (1993). However, our results agree with the most recent proposals (Quesada Pacheco 2014). The orange dialect spans the south of the United States, Mexico, Central America and the Caribbean countries whereas the green variety comprises the Rio de la Plata region in Argentina, Uruguay and Paraguay. The former dialect has been more influenced over the years by a closer relationship with the European metropolis (see also the blue dots in the Mexican plateau). In contrast, the Rioplatense Spanish has its own personality due to later settlement of the region and less contact with the prestigious norm radiating from Spain during the colonial age. It is interesting to note that the strongly mixed character of Chile suggests that this country could, in fact, build up a different region by itself, in agreement with recent schemes (Cahuzac 1980). Future work, which should increase the size of the dataset and reduce the noise levels, might detect more clearly this language variety and also shed more light on the nature of the yellow dialect, which now appears to be quite dispersed.

**Figure 5: Geographical distribution of language varieties corresponding to superdialect β. Note the separation between European and American dialects and the division of the latter into three main blocks.**

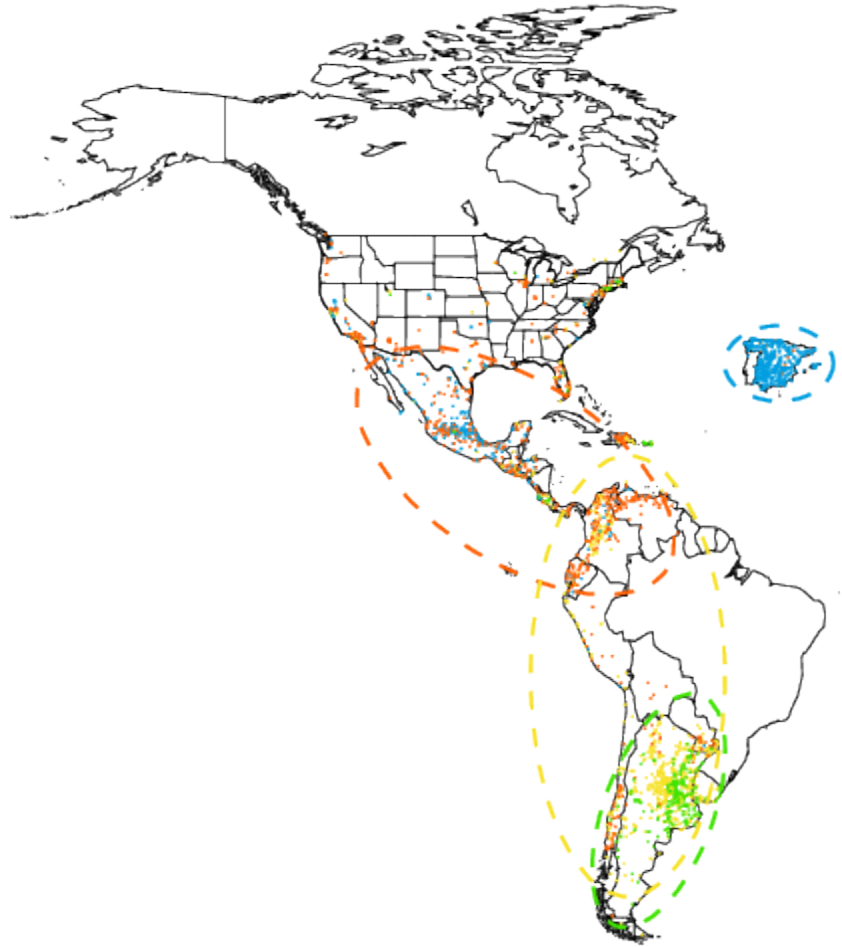

## 4. Conclusions

As technology permeates more and more aspects of our daily lives, our ability to observe human behavior through different lenses also increases. In particular, the advent of large-scale online social media services implies that we can, for the first time in history, perform a detailed analysis of how language is used informally around the world. In this manuscript we extend our previous efforts in employing user-generated content to analyze the diatopic structure of modern day Spanish language at the lexical level. Using a completely new list of concepts and related words and a much larger geolocated Twitter corpus, we recover a more detailed geographical picture of the two superdialects identified and characterize the regional variations observed. The robustness of our results points the way towards a deeper understanding of vernacular language and opens the doors to more detailed empirical studies of language usage and evolution across the world. This study is one further step towards a large-scale approach to linguistics and we believe that similar studies on other languages will undoubtedly allow us to start to glimpse common mechanisms in language evolution and differentiation.

**Appendix**

Here follows the list of concepts and words used included in our dataset. The Varilex code number (Ueda / Takagaki / Ruiz Tinoco 1993) is shown in brackets.

| **Concept** | **Lexical features** |
|---|---|
| 'Slice of cheese' [B009] | *lámina de queso, lasca de queso, loncha de queso, lonja de queso, rebanada de queso, rodaja de queso, slice de queso, tajada de queso, queso de sandwich, queso en lonchas, queso en rebanadas, queso en slice, queso americano, tranchetes* |
| 'Demijohn' [B011] | *bidón, bombona, botella grande, garrafa, garrafón, tambuche, candungo, pomo plástico* |
| 'Washer' [B038] | *lavadora, lavarropa, lavarropas, máquina de lavar* |
| 'Plaster' [B046] | *banda adhesiva, curita, esparadrapo, tirita* |
| 'Attic' [B051] | *ático, altillo, azotea, buhardilla, guardilla, penthouse, mansarda, tabanco* |
| 'Wardrobe' [B055] | *armario, closet, placard, ropero, guardarropas* |
| 'Braces, suspenders' [B077] | *breteles, brutales, suspensores, tiradores, tirantes* |
| 'Ring' [B097] | *anillo, argolla, aro, sortija, cintillo* |
| 'Tape recorder' [B111] | *cassette, casete, grabador, grabadora, magnetofón, tocacintas, magnetófono* |
| 'Blind man's buff' [B119] | *escondidas, gallina ciega, gallinita ciega, gallito ciego, pita ciega, gallo ciego* |
| 'Merry-go-round' [B134] | *caballitos, calesita, carrusel, tiovivo, machina* |
| 'Loudspeaker' [B153] | *altavoz, altoparlante, altovoz, amplificador, megáfono, parlante, magnavoz* |
| 'Flower pot' [B170] | *maceta, macetero, matera, matero, tiesto, macetera, plantera* |
| 'Fans' [C001] | *afición, aficionados, fanáticos, fanaticada, forofos, hinchada, hinchas, seguidores* |
| 'Waiter' [C014] | *camarero, barman, mesero, mesonero, mozo, camarero* |
| 'School' [C029] | *colegio, escuela, centro escolar, scuela* |
| 'Amusement' [C028] | *distracciones, diversión, entretención, entretenimiento, pasatiempo* |
| 'Stay' [C030] | *estada, estadía, estancia* |
| 'Miss' [C031] | *equivocación, error, falencia, fallo* |
| ´Cheek´ [C058] | *cachetes, carrillos, galtas, mejillas, mofletes, pómulo* |
| 'Monkey' [C060] | *chango, chimpancé, macaco, mono, mico, simio, chongo* |
| 'Mosquito' [C061] | *cínife, mosco, mosquito, zancudo* |
| 'Chance' [C065] | *bicoca, chance, ocasión, oportunidad* |
| 'Parcel, package' [C066] | *encomienda, paquete postal* |
| ´Sponsor' [C072] | *auspiciador, auspiciante, espónsor, patrocinador, patrocinante, propiciador, sponsor* |
| 'Banana' [C080] | *banana, banano, cambur, guineo, plátano, tombo* |
| 'Dust' [C082] | *nube de polvo, polvadera, polvareda, polvazal, polvero, polvoreda, polvorín, terral, terregal, tierral, tolvanera* |
| 'Bar' [C107] | *bar, boliche, cantina, cervecería, pulpería, taberna, tasca, expendio, piquera* |
| 'Earthquake' [C109] | *movimiento telúrico, movimiento sísmico, remezón, seísmo, sismo, temblor de tierra, terremoto* |
| ´Shooting' [C112] | *abaleo, balacera, baleada, tiroteo* |
| 'Glance' [C116] | *ojeada, miradita, vistazo* |
| ´Greasy' [C156] | *engrasado, grasiento, grasoso, mantecoso, seboso* |
| ´Beautiful' [C159] | *bella, bonita, hermosa, linda, preciosa* |
| ´Cold' [C182] | *catarro, constipado, coriza, gripa, gripe, resfrío, resfriado, trancazo* |
| 'Cellophane tape' [E007] | *celo, celofán, cinta adhesiva, cinta scotch, cintex, scotch, teip, dúrex, diurex, cinta pegante* |
| 'Crane' [E013] | *grúa, guinche, tecle* |
| 'Fruit cup' [E017] | *ensalada de frutas, macedonia, clericó, cóctel de frutas, tuttifruti, tutifruti* |
| 'Gas station' [E018] | *bomba de gasolina, bomba de nafta, estación de servicio, gasolinera, bencinera, bomba de bencina, gasolinería, surtidor de gasolina* |
| 'Interview' [E020] | *entrevistar, reportear, interviuvar* |
| 'Obstinate' [E026] | *cabezón, cabezudo, cabeza dura, cabezota, obstinado, porfiado, terco, testarudo, tozudo* |
| 'Peanut' [E027] | *cacahuate, cacahuete, maní, cacahué, cacaomani* |
| 'Scratch' [E032] | *arañazo, arañón, aruñetazo, aruñón, rajuño, rayón, rasguño, rasguñón* |
| 'Sweetener' [E036] | *edulcorante, endulzante, endulcina, endulzador, sacarina* |
| 'Thaw' [E039] | *descongelar, deshielar* |
| 'Miss' [F125] | *echar de menos, extrañar, añorar* |
| 'Park' [D037] | *aparcar, estacionar, parquear* |